\documentclass[11pt]{article}

\usepackage[review]{EMNLP2022}

\usepackage{times}
\usepackage{latexsym}

\usepackage[T1]{fontenc}

\usepackage[utf8]{inputenc}

\usepackage{microtype}

\usepackage{inconsolata}
\usepackage{amsfonts}
\usepackage{amsthm,amsmath,amssymb}
\usepackage{mathrsfs}
\usepackage{graphicx}
\usepackage{subfigure}
\usepackage{multirow}
\usepackage{diagbox}

\title{From Easy to Hard: Two-stage Selector and Reader \\ for Multi-hop Question Answering}

\author{Xin-Yi Li, Wei-Jun Lei, Yu-Bin Yang \\
        State Key Laboratory for Novel Software Technology \\
        Nanjing University, Nanjing 21023, China \\
        \{lixinyi,leiweijun\}@smail.nju.edu.cn,\; yangyubin@nju.edu.cn \\
        }

\begin{document}
\maketitle

\begin{abstract}
Multi-hop question answering (QA) is a challenging task requiring QA systems to perform complex reasoning over multiple documents and provide supporting facts together with the exact answer. Existing works tend to utilize graph-based reasoning and question decomposition to obtain the reasoning chain, which inevitably introduces additional complexity and cumulative error to the system. To address the above issue, we propose a simple yet effective novel framework, From Easy to Hard (FE2H), to remove distracting information and obtain better contextual representations for the multi-hop QA task. Inspired by the iterative document selection process and the progressive learning custom of humans, FE2H divides both the document selector and reader into two stages following an easy-to-hard manner. Specifically, we first select the document most relevant to the question and then utilize the question together with this document to select other pertinent documents. As for the QA phase, our reader is first trained on a single-hop QA dataset and then transferred into the multi-hop QA task. We comprehensively evaluate our model on the popular multi-hop QA benchmark HotpotQA. Experimental results demonstrate that our method ourperforms all other methods in the leaderboard of HotpotQA (distractor setting).


\end{abstract}

\section{Introduction}
\label{sec:intro}
\begin{figure}[htbp]
\centering\includegraphics[height=6.5cm]{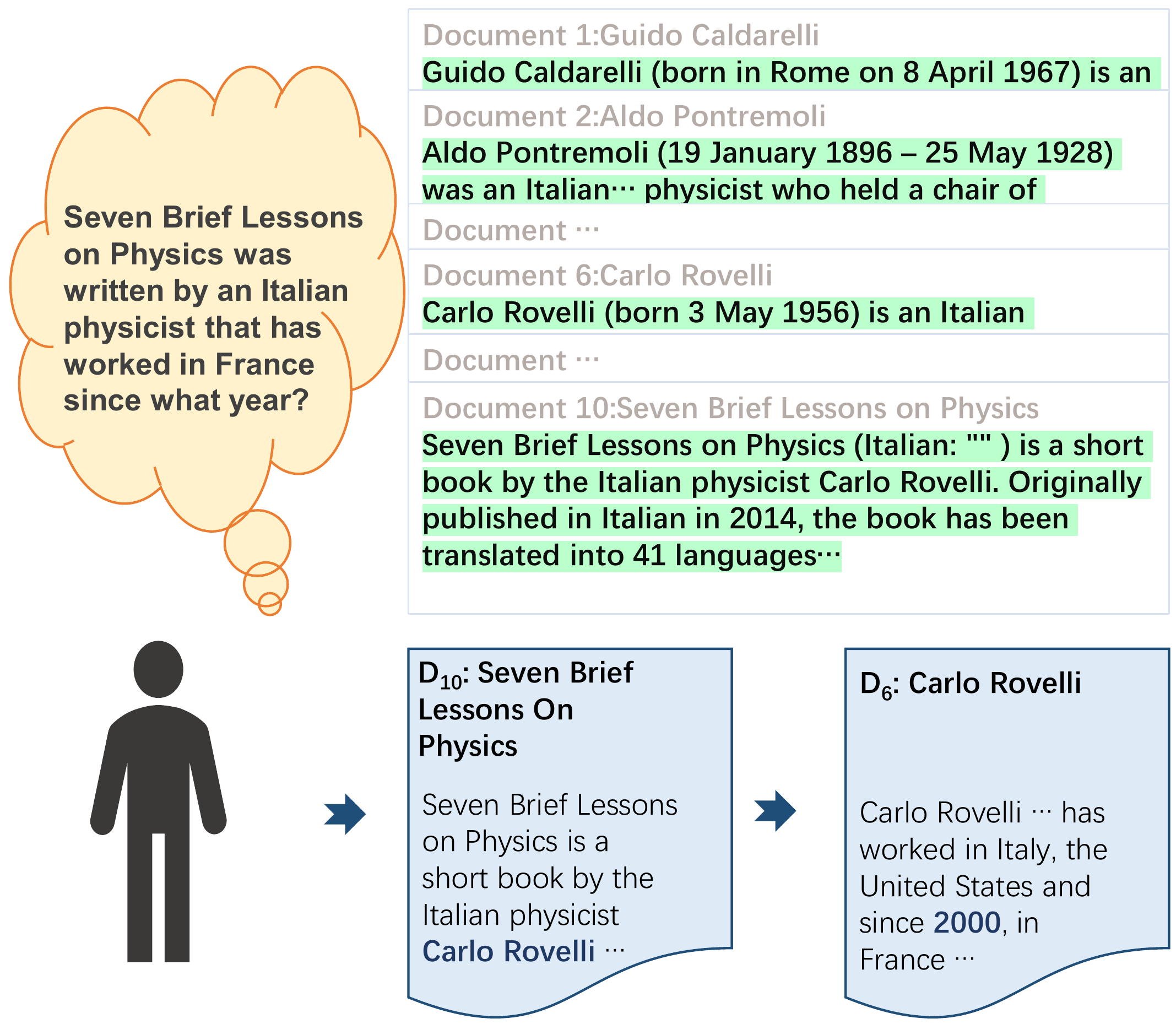}
\caption{An example of multi-hop QA from HotpotQA benchmark. The illustrated reasoning process echoes our two-stage document selection that performed from easy to hard.}
\label{fig1}
\end{figure}

Question answering (QA) is a popular benchmark task in natural language understanding that aims to teach the machine to answer questions according to the given context. Earlier QA datasets such as SQuAD \cite{rajpurkar2016squad} mainly focus on evaluating the ability of finding answers with a single document per question. Subsequently, more challenging datasets that provide each question with multiple documents appear \cite{joshi-etal-2017-triviaqa,DBLP:journals/corr/DunnSHGCC17}. However, most questions of these datasets could still be solved using only one document. Recently, many datasets \cite{welbl-etal-2018-constructing,talmor-berant-2018-web,yang-etal-2018-hotpotqa} that require multi-hop reasoning over multiple documents are constructed. In this paper, we focus on HotpotQA \cite{yang-etal-2018-hotpotqa}, which is one of the representative multi-hop QA benchmarks. Each question in the HotpotQA \cite{yang-etal-2018-hotpotqa} is provided with a set of documents that includes both relevant and irrelevant documents. 
In addition to the exact answer spans, HotpotQA \cite{yang-etal-2018-hotpotqa} also offers supporting sentences that served as explanations to train and evaluate the QA systems.

Following the current mainstream pipeline for multi-hop QA \cite{fang-etal-2020-hierarchical,Tu2019a,Wu2021}, our proposed framework also consists of a selector to minimize distracting information and a reader to provide exact answers and explanations. As for the design of the selector, most earlier works ignore the multi-hop nature of the task and treat the candidate documents or sentences independently. To address this problem, recent works include SAE \cite{Tu2019a} and S2G \cite{Wu2021} apply a multi-head self-attention (MHSA) layer to encourage inter-document interactions. However, they still use a pre-trained language model (PLM) to encode each document separately before this MHSA layer, which cannot fully exploit the strong modeling and natural language understanding capacity of the PLMs and thus results in limited interactions. 
Notice that although S2G further adopts cascaded retrieval that fed with the text of the coarse-selected documents compared to SAE, these selectors aim to find all of the relevant documents at one time, ignoring the sequential nature of the relevant documents brought by the multi-hop reasoning.


Different from the above-mentioned selectors, our proposed selector consists of two stages, where one document is selected at a stage. Specifically, we first select the document that is most relevant to the question and then select the other one based on the question and the previously selected document. Figure \ref{fig1} illustrates an example of this two-stage document selection process when we are answering multi-hop questions, which suggests that our two-stage selector shows great consistency to the reasoning process of humans. Concretely, we first arrive at document $D_{10}$ according to the question, which indicates that ``Seven Brief Lessons on Physics is written by Carlo Rovelli''. Next, we identify the document containing ``Carlo Rovelli has worked in France since 2000'' using the question together with document $D_{10}$. Our experiments demonstrate that with the help of the two-stage easy-to-hard manner, even simply adding a binary classification layer on a PLM could achieve strong performance.

After document selection, the filtered context is passed to the question answering module, namely the reader. Many existing works focus on guiding the readers to imitate the human reasoning process with elaborately designed pipelines, which can be roughly divided into two categories. One category is graph-based approaches, including SAE \cite{Tu2019a} that treats sentences as graph nodes and HGN \cite{fang-etal-2020-hierarchical} that builds a hierarchical graph with nodes of different granularities. However, these methods require additional techniques, including named entity recognition and graph modeling, which inevitably lead to the growth of complexity and error accumulation. Due to these disadvantages, there have been works demonstrating that graph modeling may not be indispensable \cite{shao-etal-2020-graph}. Consequently, we also do not exploit any graph-based reasoning methods for the sake of simplicity.
Another category is question decomposition methods, such as \textsc{DecompRC} \cite{min-etal-2019-multi} and ONUS \cite{perez-etal-2020-unsupervised} that map one multi-hop, hard question to many single-hop, easy sub-questions. Although we appreciate the idea of decomposing the hard task into easy ones, the difficulty and complexity brought by the sub-question generation cannot be ignored. 

Inspired by the question decomposition-based methods and the experience that children are usually learned from the very easy questions, we pose the question that whether it is too hard and confusing for models to directly learn to answer multi-hop questions from scratch. Therefore, our reader is firstly trained on a simpler single-hop dataset SQuAD \cite{rajpurkar-etal-2016-squad} before performing the harder, multi-hop QA task. We assume that training on the single-hop dataset could initialize the model with better parameters and make it easier for the model to answer multi-hop questions. 
In addition, the process of learning from easy to hard is consistent with the widely accepted concept that there supposed to be a gradual progression of skill acquisition \cite{liu2017easy,roads2018easy,song2019easy}. Our experiments illustrate that the two-stage reader effectively boosts the performance, which verifies our assumption.

The main contributions of our work are summarized as follows:

\begin{itemize}
    \item We propose a simple yet effective multi-hop QA framework called FE2H, where both the document selector and reader are divided into two stages following the easy-to-hard manner. 
    \item We introduce a novel document selection module that iteratively performs binary classification tasks to select relevant documents by simply adding a prediction layer on a PLM.
    \item The proposed reader is first trained on a single-hop QA dataset and then transferred into the multi-hop QA task, which is inspired by the progressive learning process of humans.
    \item We comprehensively evaluate our model on the popular multi-hop QA benchmark HotpotQA. Experimental results show that the proposed FE2H can surpass the previous state-of-the-art by a large margin across all metrics.
\end{itemize}

\section{Related Work}
\label{sec:related}
\paragraph{Selector for QA}
Sentence or document selector aims to minimize noisy information, which has been proved effective for improving the performance and efficiency of the QA system. The study in \cite{min-etal-2018-efficient} proposes a sentence selector for SQuAD \cite{rajpurkar-etal-2016-squad} since questions can be answered within a single sentence. As for multi-hop QA, \cite{wang-etal-2019-multi} and \cite{groeneveld-etal-2020-simple} simply treat the documents and sentences separately, ignoring the interaction between them. Therefore, SAE \cite{Tu2019a} and S2G \cite{Wu2021} append an MHSA layer to encourage document interaction. To acquire the finer-grained result, S2G \cite{Wu2021} further adopt another cascaded paragraph retrieval module, which takes the text of the selected documents as input and subsequently explore deeper relationships between them. Since SAE and S2G attempt to locate all of the relevant documents at the same time, a naive binary classifier could not perform well. Therefore, SAE \cite{Tu2019a} and S2G \cite{Wu2021} reformulate the classification objective into a ranking and scoring objective respectively to conform with the ranking nature of the selector. Different from the above selectors, we present a novel two-stage selector that neither selects the documents independently nor simultaneously.

\paragraph{Reader for multi-hop QA} Existing multi-hop QA readers can be roughly divided into two categories: reasoning based on intermediate results and reasoning based on graph. The first category can be further grouped by whether the intermediate result is represented in hidden state or free form text. As for the former group, QFE \cite{nishida-etal-2019-answering} identifies relevant sentences by updating an RNN hidden state vector in each step, others update the representation vector of query and context iteratively to obtain the final answer \cite{das2018multi,feldman-el-yaniv-2019-multi}. The representative works of the latter group mainly acquire the intermediate textual result by firstly decomposing the original question or generating a sub-question and then adopting off-the-shelf readers \cite{qi-etal-2019-answering,min-etal-2019-multi,perez-etal-2020-unsupervised}. The second category exploits graph neural networks to reason over the constructed graph. DFGN \cite{qiu-etal-2019-dynamically} and CogQA \cite{ding-etal-2019-cognitive} focus on the entity graph while SAE \cite{Tu2019a} treats sentences as graph nodes to identify supporting sentences. Furthermore, \cite{tu-etal-2019-multi}, HGN \cite{fang-etal-2020-hierarchical} and AMGN \cite{AMGN+:Li2021} reasoning over the heterogeneous graph with different types of nodes and edges. However, there are recent works that question whether the graph structure is necessary. \textsc{Quark} \cite{groeneveld-etal-2020-simple}, C2F Reader \cite{shao-etal-2020-graph} and S2G \cite{Wu2021} demonstrate that the performances of the simple graph-free readers are on par with the graph-based reasoning methods. C2F Reader \cite{shao-etal-2020-graph} further claims that graph-attention can be considered as a special case of self-attention. In this paper, we propose a reader that simply follows a two-stage pipeline and free of any recurrent state updating and graph modeling.

\section{Proposed Framework}
\begin{figure*}[htbp]
\centering\includegraphics[height=5.5cm]{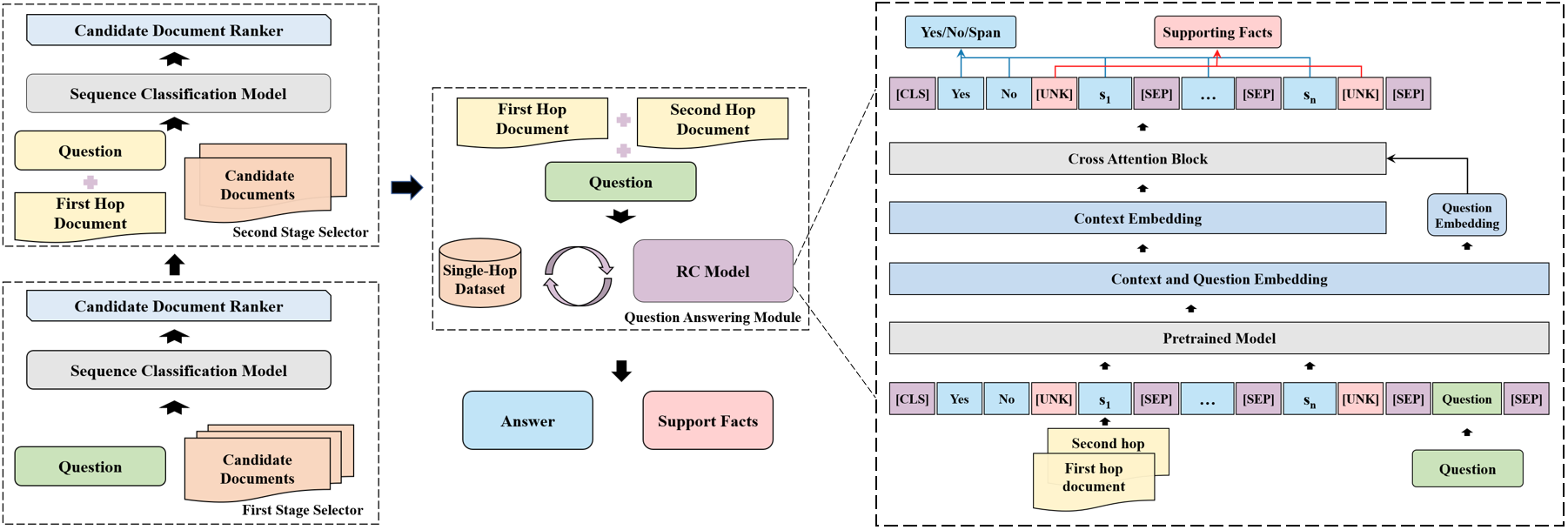}
\caption{Overview of our proposed framework FE2H. The left part shows our whole pipeline, including a first stage selector to find the most relevant document to the question, a second stage selector to find the other relevant document according to the question and the previously selected document, and a question answering module to extract answers and supporting facts based on the selected documents and question. The right part illustrates the detailed computation process of our cross attention enhanced reading comprehension model.}
\label{fig2}
\end{figure*}
We choose the distractor setting of HotpotQA \cite{yang-etal-2018-hotpotqa} as our testbed, where each question is equipped with 2 relevant documents and 8 distractors. A list of supporting sentences and an answer that could be an exact text span or “Yes/No” are also provided for evaluation. The overview of our proposed framework is shown in Figure \ref{fig2}. Following the traditional pipeline of multi-hop QA system, We first utilize a document selection module to construct almost noise-free context and then pass them to the downstream question answering module to extract supporting facts and answer span simultaneously.

\subsection{Document Selection Module}

At this phase, we aim to filter the distracting information and generate high-quality context, namely almost noise-free context, for the following question answering module. Note that although there are many works that perform recurrent reasoning using graph modeling and question decomposition when extracting supporting sentences and answers, existing selectors still extract relevant documents independently or simultaneously, which has been discussed in Section \ref{sec:intro} and Section \ref{sec:related}. Therefore, we assume that the recurrent reasoning manner should also be applied in a larger granularity, namely the document level. Recall that when we answer a multi-hop question, in most cases, we first select the document most relevant to the question and use this document together with the question to find another. Inspired by this insight, we divide the document selection into two steps, one document at a step, which is different from the existing methods. 

\paragraph{The first stage}In this stage, we aim to select the document that is most relevant to the question. We feed the sequence ``[CLS] + question + [SEP] + document + [SEP]'' to a PLM ELECTRA \cite{Clark2020} and project the output of token ``[CLS]'' to calculate the relevant score $P(d|q)$ for each candidate document $d$ and question $q$. We label all the documents containing supporting facts as relevant and calculate the binary cross-entropy loss as:
\begin{equation}
    \begin{aligned}
	\mathcal{L}_{S_1}=-\sum_{i=1}^{N}\sum_{j=1}^{M}(t_{ij}\log P(d_{ij}|q_i)\\+(1-t_{ij})\log(1-P(d_{ij}|q_i)))
    \end{aligned}
\end{equation}
where $q_i$ is the $i$th question in the dataset, $d_{ij}$ is the $j$th candidate document of $q_i$, $t_{ij}$ is the label of the question-document pair $(q_i,d_{ij})$, $N$ is the number of questions and $M$ is the size of each candidate document set, which is 10 for HotpotQA. We select the document with the highest relevance score and denote the chosen document as $p_1$. That is, given question $q_i$, $p_{i1}=\mathop{\arg\max}\limits_{j}P(d_{ij}|q_i)$.

\paragraph{The second stage}The aim of this stage is to find the other relevant document based on the previous result. Therefore, we generate the input as ``[CLS] + question + [SEP] + $\rm{document}_1$ + [SEP] + $\rm{document}_2$ + [SEP]'', where $\rm{document}_1$ is the document selected at the first step and $\rm{document}_2$ is another candidate document. We calculate the relevance score of the candidate document as the previous step, which is denoted as $P(d|q,p)$. The binary cross-entropy loss is computed as:
\begin{equation}
    \begin{aligned}
	\mathcal{L}_{S_2}=-\sum_{i=1}^{N}\sum_{j=1}^{M}\mathbb I(d_{ij}\neq p_{i1})  (t_{ij}\log P(d_{ij}|q_i,p_{i1})\\+(1-t_{ij})\log(1-P(d_{ij}|q_i,p_{i1})))
    \end{aligned}
\end{equation}

where $\mathbb I$  is the indicator function. Likewise, we select the document with the highest relevance score of this step and denote it as $p_{2}$. 

After the two-stage retrieval, the selected documents for each question are concatenated and passed to the following question answering module to serve as the high-quality context.

\subsection{Question Answering Module}
\paragraph{Multi-task Model}
In this phase, we implement a multi-task model to extract the answers and the supporting facts of the multi-hop questions, which is illustrated in Figure \ref{fig2}. Firstly, we concatenate the context and the question as a sequence and use a PLM called ELECTRA \cite{Clark2020} to obtain the contextual representations of the input text. Subsequently, We exploit cross attention to enhance the interaction between the context and the question to acquire better context embeddings. Finally, a multi-task learning approach is adopted to jointly extract the answer and supporting sentences by simply adding a linear prediction layer.

Firstly, we are supposed to construct the input sequence as the input for ELECTRA \cite{Clark2020}. Considering the ``Yes/No'' answer case, ``yes'' and ``no'' tokens are added to the beginning of the sequence. Besides, we add a ``[UNK]'' before each sentence in the context, whose contextual representation is used for supporting sentence classification. Therefore, our input sequence can be formulated as ``[CLS] + yes + no + context + [SEP] + question + [SEP]''. We denote the ELECTRA output of the input as $H=[{{h}_{1}},\cdots ,{{h}_{L}}]\in {{\mathbb{R}}^{L\times {d}}}$ , the context output as ${{H}_{c}}=[{{h}_{1}},\cdots ,{{h}_{S}}]\in {{\mathbb{R}}^{S\times {d}}}$ and the question output as ${{H}_{q}}=[{{h}_{S+1}},\cdots ,{{h}_{L}}]\in {{\mathbb{R}}^{T\times {d}}}$ , where $L$ is the length of the input sequence, $S$ is the length of the context, $T=L-S$ is the length of the query, and ${d}$ is the hidden vector dimension of ELECTRA.

To improve the QA performance, we further add a cross attention interaction module to enhance the interaction between the context and query representations. At the same time, in order to obtain a better result and make the convergence faster during training, we use two cross attention blocks, including one with layer normalization and the other without layer normalization. The cross attention is computed as:
\begin{align}
   {Q}&=H_{c}W^{Q}\! \\ 
   {K}&=H_{q}W^{K}\! \\ 
   {V}&=H_{q}W^{V}\! \\ 
   \operatorname{CrossAttention}&=\operatorname{Attention}(Q,K,V) \\
   \operatorname{Attention}(Q,K,V)&=\operatorname{softmax}(\frac{Q{{K}^\mathsf{T}}}{\sqrt{{{d}_{k}}}})V
\end{align}

To extract the answer spans, we utilize a linear prediction layer on the context representations to identify the start and end position of the answer. The corresponding loss items are denoted as ${\mathcal{L}_{start}}$ and ${\mathcal{L}_{end}}$ respectively. As for the supporting facts prediction, we utilize a linear binary classifier on the output of the ``[UNK]'' that added before each sentence to decide whether a sentence is relevant or not. The classification objective of supporting facts is denoted as ${\mathcal{L}_{sf}}$. Finally, we jointly optimize the above objectives as:
\begin{align}\label{qa-loss}
    \mathcal{L}={{\gamma }_{1}}{\mathcal{L}_{start}}+{{\gamma }_{2}}{\mathcal{L}_{end}}+{{\gamma }_{3}}{\mathcal{L}_{sf}}
\end{align}

\paragraph{Two-stage Reader}
As discussed in Section \ref{sec:intro}, we assume that directly fine-tuned a PLM on a multi-hop QA task may be too hard and confusing for models. As a result, the model may answer the multi-hop questions with the single-hop shortcut, namely word-matching the question with a single sentence \cite{jiang-bansal-2019-avoiding}. Therefore, we first train our model on a single-hop QA dataset SQuAD \cite{rajpurkar-etal-2016-squad} and then transfer the model to the multi-hop QA task. We suppose that the parameters that optimized by the single-hop QA task are much better than the random initialized parameters. Consequently, after learning single-hop QA task, the model tends to converge faster and achieve higher performance. Note that the model structure of the single-hop and multi-hop stage is the same.
\section{Experiments}
\begin{table*}[htbp]
\centering
\begin{tabular}{lllllll}
\hline
\multicolumn{1}{c}{\multirow{2}{*}{Model}} & \multicolumn{2}{c}{Ans}                               & \multicolumn{2}{c}{Sup}                            & \multicolumn{2}{c}{Joint}                       \\ \cline{2-7} 
\multicolumn{1}{c}{}                       & \multicolumn{1}{c}{EM}    & \multicolumn{1}{c}{F1}    & \multicolumn{1}{c}{EM}    & \multicolumn{1}{c}{F1} & \multicolumn{1}{c}{EM} & \multicolumn{1}{c}{F1} \\ \hline
Baseline Model \cite{yang-etal-2018-hotpotqa}  & \multicolumn{1}{c}{45.60} & \multicolumn{1}{c}{59.02} & \multicolumn{1}{c}{20.32} & 64.49& 10.83& 40.16\\
QFE \cite{nishida-etal-2019-answering}& \multicolumn{1}{c}{53.86} & \multicolumn{1}{c}{68.06} & \multicolumn{1}{c}{57.75} & 84.49& 34.63& 59.61\\
DFGN \cite{qiu-etal-2019-dynamically} & \multicolumn{1}{c}{56.31} & \multicolumn{1}{c}{69.69} & \multicolumn{1}{c}{51.50} & 81.62& 33.62&59.82 \\
SAE-large \cite{Tu2019a}  & 66.92 & 79.62 & 61.53 & 86.86 & 45.36 & 71.45\\
C2F Reader \cite{shao-etal-2020-graph} & 67.98 & 81.24 & 60.81  & 87.63 & 44.67 & 72.73\\
HGN-large \cite{fang-etal-2020-hierarchical}  & 69.22 & 82.19 & 62.76 & 88.47 & 47.11 & 74.21 \\ 
AMGN+ \cite{AMGN+:Li2021}  & 70.53 & 83.37 & 63.57            & 88.83 & 47.77 & 75.24 \\
S2G+EGA \cite{Wu2021}  & 70.92 & 83.44 & 63.86            & 88.68 & 48.76 & 75.47 \\
SAE+\textsuperscript{\ref{hotpotqa_url}}  & 70.74	 & 83.61 & 63.70            & 88.95 & 48.15 & 75.72 \\
\hline
FE2H on ELECTRA(ours)& 69.54 & 82.69 & 64.78& 88.71 & 48.46 & 74.90\\
FE2H on ALBERT(ours)& \textbf{71.89} & \textbf{84.44} & \textbf{64.98	}& \textbf{89.14	} & \textbf{50.04} & \textbf{76.54}\\ \hline
\end{tabular}
\caption{\label{leaderboard} Comparison of the results on the test set of HotpotQA in the distractor setting. FE2H model obtains the overall competitive performance and achieves the state-of-the-art results in all metrics on the leaderboard among all works including unpublished methods such as SAE+\textsuperscript{\ref{hotpotqa_url}}, demonstrating the strong multi-hop reasoning ability of our model.}
\end{table*}
\subsection{Dataset}

We evaluate our model on the distractor setting of HotpotQA \cite{yang-etal-2018-hotpotqa}. HotpotQA consists of 113k Wikipedia-based and crowdsourced question-answer pairs, including 90K training examples, 7.4K development examples and 7.4K test examples. 

In the distractor setting, each question in the dataset is provided with a small set of documents that contains two supporting documents and eight irrelevant documents. HotpotQA \cite{yang-etal-2018-hotpotqa} also provides a list of supporting sentences for each question, which encourages model to explain the predictions. The Exact Match (EM) and F1 scores are used to evaluate models for both tasks and joint EM and F1 scores are used in the evaluation of the overall performance. The performance of our model on the test set is obtained by submitting our best model on the development set. Since the test set is not available, we report the performance of our comprehensive experiments on the development set.

\subsection{Implementation Details}
Our model is implemented based on the transformers of Hugging Face \cite{wolf-etal-2020-transformers} and we adopt ELECTRA and ALBERT as our PLM to obtain the contextual embeddings of the questions and context. Our experiments are all performed on 2 or 3 NVIDIA A100 GPUs. The implementation details of our two-stage selector and reader are listed as follows. 
\paragraph{Two-stage Selector}
We train two document selectors with a base and a large version of ELECTRA, respectively. Due to resource limitation, the ablations on document selection is only performed on ELECTRA-base. The number of epochs is set to 3 and batch size is set to 12. We use BERT-Adam with learning rate of 2e-5 for the optimization. The hyper-parameters of our two selectors of different stages are same while the model parameters are different and trained sequentially.
\paragraph{Two-stage Reader}
As for the question answering phase, we train the FE2H on ELECTRA model with batch size of 16 with 2 NVIDIA A100 GPUs, learning rate of 1.5e-5, warm-up rate of 0.1, and L2 weight decay of 0.01. Besides, we use ELECTRA-large as the PLM of this phase. It takes about 2 hours per epoch and 6.5 hours in total to train a reader of a single stage. we train the FE2H on ALBERT model with batch size of 12 with 3 NVIDIA A100 GPUs, learning rate of 1.0e-5, warm-up rate of 0.1, and L2 weight decay of 0.01. Besides, we use Albert-xxlarge-v2 as the PLM of this phase. It takes about 25 hours in total to train a reader of a single stage.

\subsection{Experimental Results}

\paragraph{Results on HotpotQA}
Table \ref{leaderboard} shows the performance of published advanced methods on the test set of HotpotQA in the distractor setting. Despite the simplicity of our model, Our FE2H on ALBERT is the state-of-the-art model on the leaderboard\footnote{Leaderboard available at \url{https://hotpotqa.github.io}\label{hotpotqa_url}}. Our model achieves the best result in all metrics, demonstrating the strong multi-hop reasoning ability of our model. The Joint EM improvement is 1.28\%  and the Joint F1 improvement is 1.07\% among all the published methods. And the Joint F1 improvement is 0.82\% among all the submitted methods including SAE+\textsuperscript{\ref{hotpotqa_url}}.
\begin{table}[htbp]
\centering
\begin{tabular}{lcc}
\hline
Model & EM & F1 \\ \hline
$\mathrm{SAE}_{large}$ &91.98 & 95.76 \\
HGN         &- & 94.53 \\
$\mathrm{S2G}_{large}$ & \underline{95.77} &\underline{97.82} \\
$\mathrm{FE2H}_{base}$ (Ours)  & 95.53     & 97.59   \\
$\mathrm{FE2H}_{large}$ (Ours)  & \textbf{96.32}     & \textbf{98.02}   \\
\hline
\end{tabular}
\caption{\label{results-two-stage-selector} Document selection results on the dev set of HotpotQA. ``-'' means the data is not available.Bold ones indicate state of the art, and underlined ones indicate the second best result.}
\end{table}
\paragraph{Document Selection}
We compare our selector with the ones of three published advanced works, that is, HGN, SAE and S2G, which also apply carefully designed selectors to retrieve relevant documents for the downstream readers. We use EM and F1 to evaluate the performance of the document selectors. As shown in Table \ref{results-two-stage-selector}, our selector outperforms these three strong baselines. Even the base version of our selector achieves competitive performance compared to the previous best selector.

\subsection{Ablation}

\begin{table}[htbp]
\centering
\begin{tabular}{lcc}
\hline
Setting & EM & F1 \\ \hline
FE2H  & \textbf{95.53}     & \textbf{97.59}   \\
1st Stage + Top 2  & 89.10    & 94.41    \\
1st Stage + THOLD   & 85.10    & 87.01   \\ 
1st Stage (AW) + Top 2& 84.17    & 91.78  \\
1st Stage (AW) + 2nd Stage & \underline{92.72}    & \underline{96.09}    \\\hline

\end{tabular}
\caption{\label{ablation-two-stage-selector} Ablation on two-stage selector on dev set of HotpotQA. works. ``THOLD'' means using a threshold to select documents. ``AW'' means training with assigned weights.}
\end{table}
\paragraph{Two-stage Selector Ablation}
The results of the two-stage selector ablation are shown in Table \ref{ablation-two-stage-selector}. 
To prove the validity of out two-stage selector, we remove the second stage and evaluate the results. Results illustrate that, given the output of the first stage, no matter we select the top 2 documents or select the documents whose scores above a certain threshold, both EM and F1 score drop significantly. Note that both SAE and S2G believe that the documents containing the answer span (i.e., gold documents) are more important for downstream tasks and thus assign these documents with higher score. However, we assume that relevant documents are of equal importance,  especially for the first stage of our document selection. Since we do not reformulate the binary classification objective to ranking-related objectives like SAE and S2G, we assign greater weights to the gold documents to verify this assumption. Results illustrate that performance drops significantly after reassigning the weights of the relevant documents as above at the first stage. It is worth pointing that the performance of single-stage setting degrades more compared to the two-stage setting, indicating that directly guiding the selector to identify the gold documents is infeasible, thus both relevant documents are equally important.

\begin{table}[htbp]
\centering
\scalebox{0.9}{
\begin{tabular}{lcc}
\hline
Setting                      & Joint F1 & Joint EM \\ \hline
Ours  & \textbf{75.30}    & \textbf{48.55}    \\
\quad- Cross Attention w/ LN    & 75.14    & 48.37    \\
\quad- Cross Attention w/o LN &  75.07   & 48.37    \\
\quad- Cross Attention Block & 75.20 & 48.19 \\
Two Cross Attention w/ LN  &  74.98    & 48.31    \\
Two Cross Attention w/o LN  &75.18    & 48.29   
\\\hline
Feed-Forward Network       & 75.05    & 48.00    \\
Self-Attention               & 74.85    & 48.00    \\ 
Bi-Attention               & 74.47    & 47.29    \\
Co-Attention               & 74.10    & 46.92    \\
\hline
\end{tabular}
}
\caption{\label{ablation-qa-module}Ablation results on components of the question answering module.}
\end{table}

\paragraph{QA Model Ablation}
Due to the computing resource limitations, we only use a single-stage reader, i.e. the reader without training on the single-hop QA dataset, to perform the ablation of QA model structure, which is shown in Table \ref{ablation-qa-module}. Specifically, we investigate the effect of different interaction layers between the question and the context on the performance of extracting supporting facts and answers.

Recall that our QA model exploits a cross attention block that consists of a cross attention layer with layer normalization and one without. We first ablate the cross attention block of our QA model. As shown in Table \ref{ablation-qa-module}, both the joint F1 score and joint EM drops whether one or two cross attention layers are removed. Furthermore, in order to explore the effect of the asymmetric structure of our cross attention block, we replace it with two same cross attention layers with and without layer normalization, respectively. Experimental results show that our cross attention block outperforms the above two settings. We hypothesis that this is because the one with layer normalization tends to converge faster \cite{ba2016layer} and the one without layer normalization tends to explore a better global solution due to the randomness of the layer. As a result, our cross attention block combines the advantages of both and achieve better performance.

In addition to the cross attention layer, we attempt to explore the effect of other layers on enhancing the interaction between the question and the context, which includes self-attention, co-attention, bi-attention, and feed-forward layers. These layers are directly added  to the PLM during experiments. Note that our QA model without the cross attention block is equivalent to directly predicting results based on the output of PLM, namely the baseline QA model, while our experiments show that adding these layers all lead to performance degradation compared to the baseline. For these degradations, our intuitive explanations are listed as follows. Since the feed-forward network only takes the context embeddings as input, thus prohibits the interaction between the question and context, which probably results in performance degradation. As for self-attention, recall that the PLM has already contained many self-attention layers, thus learning an extra self-attention layer from scratch may not help. Moreover, because the interactions modeled by bi-attention \cite{seo2016bidirectional} and co-attention \cite{xiong2016dynamic} are more complicated compared to the ones of self-attention and cross-attention, we suppose that it is hard to train a model with many pre-trained self-attention layers and a random-initialized bi-attention or co-attention layer.

In summary, however, we come to the conclusion that though our cross attention block improves the performance, the improvement is not significant and many other interaction mechanisms even negatively affect the performance. That is to say, without any additional computations, directly fine-tuned PLMs could achieve competitive performance on the downstream tasks.

\begin{table}[htbp]
\centering
\scalebox{0.7}{
\begin{tabular}{ccccc}
\hline
PLM & First Stage & Second Stage & Joint F1 & Joint EM \\ \hline
BERT  & SHD & MHD & \textbf{68.58}    & \textbf{41.90}    \\
\; & MHD & - & 67.13    & 39.86    \\
\; & SHD+MHD & -        & 68.37    & 41.51    \\
\; & SHD & SHD+MHD        & 68.23    & 41.89   \\ \hline
ELECTRA  & SHD & MHD & \textbf{75.73}    & \textbf{49.37}    \\
\; & MHD & - & 75.30    & 48.55    \\
\; & SHD+MHD & - &75.43    & 48.87    \\
\; & SHD & SHD+MHD & 75.55    & 48.48    \\ \hline
\end{tabular}
}
\caption{\label{ablation-two-stage-reader}Ablation results on training strategies of the two-stage reader. SHD means single-hop dataset SQuAD and MHD means multi-hop dataset HotpotQA.}
\end{table}

\paragraph{Two-stage Reader Ablation}
To demonstrate the effectiveness of the two-stage manner of our reader, we perform our ablation studies using a smaller PLM BERT-base \cite{devlin-etal-2019-bert} and a larger PLM ELECTRA-large \cite{Clark2020}. The results are illustrated in Table \ref{ablation-two-stage-reader}, which demonstrate that the two-stage manner brings more significant performance improvement on BERT \cite{devlin-etal-2019-bert} than ELECTRA. Specifically, joint F1 drops 1.45\% and joint EM drops 2.04\% when only training on the multi-hop QA dataset. We hypothesis that this is because larger PLM like ELECTRA has learned much more common knowledge and obtained stronger natural language understanding ability, thus being further fine-tuned on simple tasks like single-hop QA may be of limited help. While as for smaller PLM like BERT \cite{devlin-etal-2019-bert}, single-hop QA datasets could offer more task-specific knowledge to the PLM and effectively bridge the gap between the PLM and the multi-hop QA task. Therefore, we recommend the two-stage reader, especially under the circumstances that resources are limited and only small PLMs are permitted.

Subsequently, we perform experiments using both single-hop and multi-hop QA datasets but with different training strategies. As illustrated in Table  \ref{ablation-two-stage-reader}, performance drops when we perform single-hop and multi-hop QA tasks simultaneously, which indicates that the performance improvement of our two-stage reader is not only brought by the use of more labeled QA pairs, but also brought by the easy-to-hard two-stage manner. Notice that only using the single-hop dataset at the first stage and using both datasets at the second stage also leads to performance degradation. We presume that this is because once the model has learned the easy task, feeding it with the datasets of both tasks in the next step is of little use and even probably negatively impact the training results.

\section{Conclusion}
We propose FE2H, a simple yet effective framework for multi-hop QA that divides both the document selection and question answering into two stages following an easy-to-hard manner. Experimental results demonstrate that since we cannot feed all of the candidate documents to the PLMs at a time due to the input length limitation, taking the multi-hop reasoning nature into consideration at the document selection phase significantly improves the overall performance. As for the subsequent QA phase, thanks to the great natural language understanding ability of the PLMs, the performance of our simple two-stage reader is better than the state-of-the-art approaches without any graph structure and explicit reasoning chains. We hope this work could facilitate more simple yet powerful multi-hop QA approaches with the help of the advanced PLMs.

\section*{Acknowledgements}
We thank many colleagues at Nanjing University for their help, particularly Qing-Long Zhang, for useful discussion.

\bibliography{anthology,custom}
\bibliographystyle{acl_natbib}

\end{document}